\documentclass{ecai} 

\usepackage{latexsym}
\usepackage{amssymb}
\usepackage{amsmath}
\usepackage{amsthm}
\usepackage{booktabs}
\usepackage{enumitem}
\usepackage{graphicx}
\usepackage{color}

\usepackage{multirow}
\usepackage{subfigure}
\usepackage{rotating} 
\usepackage{tabularx} 
\usepackage{makecell}
\usepackage[linesnumbered, ruled, vlined, noend]{algorithm2e}
\usepackage{float}

\newtheorem{theorem}{Theorem}

\newtheorem{proposition}[theorem]{Proposition}

\newcommand{\BibTeX}{B\kern-.05em{\sc i\kern-.025em b}\kern-.08em\TeX}

\begin{document}


\begin{frontmatter}


\paperid{7273} 


\title{Boosting global time series forecasting models: a two-stage modelling framework}


\author[A]{\fnms{Junru}~\snm{Ren}}
\author[A]{\fnms{Shaomin}~\snm{Wu}\thanks{Corresponding Author. Email: s.m.wu@kent.ac.uk.}}

\address[A]{Kent Business School, University of Kent, Canterbury, Kent, CT2 7FS, UK}


\begin{abstract}
	A time series forecasting model---which is typically built on a single time series---is known as a local time series model (tsLM). In contrast, a forecasting model trained on multiple time series is referred to as a global time series model (tsGM). tsGMs can enhance forecasting accuracy and improve generalisation by learning cross-series information. As such, developing tsGMs has become a prominent research focus within the time series forecasting community. However, the benefits of tsGMs may not always be realised if the given set of time series is heterogeneous. While increasing model complexity can help tsGMs adapt to such a set of data, it can also increase the risk of overfitting and forecasting error. Additionally, the definition of homogeneity remains ambiguous in the literature. To address these challenges, this paper explores how to define data heterogeneity and proposes a two-stage modelling framework: At stage one, a tsGM is learnt to identify homogeneous patterns; and at stage two,  tsLMs (e.g., ARIMA) or sub-tsGMs tailored to different groups are learnt to capture the heterogeneity. Numerical experiments on four open datasets demonstrate that the proposed approach significantly outperforms six state-of-the-art models. These results highlight its effectiveness in unlocking the full potential of global forecasting models for heterogeneous datasets.
\end{abstract}

\end{frontmatter}


\section{Introduction}
Accurate time series forecasting facilitates data-driven decision-making in business, economy, and other industries \citep{petropoulos2022forecasting}. Recently, global time series models (tsGMs), estimated on a set of time series with similar patterns, have been proposed \citep{salinas2020deepar}. Research on tsGMs is becoming more burgeoning in the time series forecasting community, compared with traditional local time series models (tsLMs) that are estimated on individual time series. By leveraging cross-series information, tsGMs effectively reduce generalisation errors and perform exceptionally well, especially for the cases where individual time series are too short to build a reliable model \citep{montero2021principles,Semenoglou2021}.

The improved performance of a tsGM relies on the assumption that the time series the model learnt from are homogeneous, namely, they share a similar pattern. However, different time series may have heterogeneity, which may stem from the differences in seasonal patterns, trends, or underlying data structures \citep{Neubauer2024}. Intuitively, it is challenging to build a well-performed tsGM on a set of different time series that contain heterogeneity. \citet{wellens2023and} and \citet{hewamalage2022global} found that the forecasting performance of tsGMs is closely related to both the degree of the homogeneity of the time series and the complexity of tsGMs, and that modelling methods such as recurrent neural networks (RNNs) and light gradient boosting models (LGBMs) can handle heterogeneity better than linear time series models. A sophisticated tsGM is required to accurately describe multiple highly heterogeneous time series simultaneously. However, such complex models are prone to overfitting, especially for the cases where the volume of the available data is limited. In summary, the forecasting performance of a tsGM is constrained by the level of heterogeneity of time series on which the tsGM is learnt, while constructing tsLMs on each individual time series completely overlooks the shared information among different time series \citep{Neubauer2024}. This further prompts some new questions: {\it how to determine the heterogeneity level of a given set of  time series and how to fully leverage the strength of tsGMs learnt on heterogeneous time series?}

To improve the capability of capturing the respective characteristics of the time series a tsGM learnt on, the divide-and-conquer methodology is employed in the literature. Clustering techniques, such as distance-based clustering \citep{Godahewa2021} and feature-based clustering \citep{Bandara2020}, are applied to divide the entire dataset into several sub-groups. Time series in a sub-group is regarded as homogeneous, and then a tsGM is learned on each sub-group. Additionally, some research considers combining local and global components, resulting in the proposal of hybrid models. \citet{smyl2020hybrid} used exponential smoothing methods to capture the local level and seasonality of each series, and then a long short-term memory (LSTM) network was globally estimated on the remaining parts of the series after removing the obtained local characteristics. Their aim is to extract and separate non-homogeneous local patterns and homogeneous global patterns. 

As evident from the above literature review, the existing methods analyse some characteristics of a set of time series to obtain clusters or remove individual-specific patterns---which can be referred to as data pre-processing---then they build forecasting models. A drawback of these methods is that the data pre-processing stage only considers characteristics of the set of time series but ignores the characteristics of modelling methods. Modelling methods play an important role as different modelling methods can capture different characteristics in the set of time series. To overcome this drawback, this paper proposes to boost forecasting models with a two-stage modelling framework: At stage one, we learn a tsGM on the entire set of time series to capture the homogeneous patterns. At stage two, we perform a cluster analysis on the residual time series---which are  calculated from stage one and contain heterogeneity and noises---and then build either tsLMs (e.g., the autoregressive integrated moving average (ARIMA) models) or tsGMs  to different clusters to capture heterogeneity.  

The novelty and contribution of this paper includes: (i) It proposes of a two-stage modelling framework to first capture the homogeneity of a given set of time series, and then the heterogeneity in the residuals is separated out; (ii) the degree of heterogeneity is quantified by considering characteristics of both the data and the learnt tsGM; (iii) it presents three propositions relating to the proposed two-stage modelling framework to provide a theoretical guarantee.

The remainder of this paper is structured as follows. Section \ref{related-work} reviews related work. Section \ref{hybrid-method} proposes a two-stage modelling framework. Section \ref{experiements} compares the cumulative errors of the proposed approach with six other models on four open datasets. Section \ref{discussion} discusses relevant issues. Section \ref{conclusion} concludes the work.
\section{Related Work}\label{related-work}
\paragraph{Clustering-based models.}
The clustering-and-then-model method is a commonly-used method in dealing with heterogeneous time series forecasting. \citet{Bandara2020} and \citet{Semenoglou2021} used $k$-means algorithms and series features on trends, seasonality, and autocorrelation to conduct feature-based clustering, and then trained a cluster-specific model on each cluster. \citet{Godahewa2021} investigated feature-based clustering, distance-based clustering, and random clustering, where dynamic time warping (DTW) distances were used, and trained multiple tsGMs on each cluster of the time series, and then constructed an ensemble model to generate final forecasts. \citet{Chen2024} constructed an adaptable channel clustering module that can be plugged into neural networks to realise Euclidean distance-based clustering using radial basis function kernels. The module consists of a cluster assigner and a cluster-aware feed forward module. \citet{Froehwirth-Schnatter2008} utilised a model-based clustering method and assumed the time series were originated from AR processes. A time series forecasting model was then integrated into the clustering. On the contrary, \citet{Neubauer2024} considered a model-and-then-clustering mechanism and proposed an algorithm named TSAVG. In their work, specifically, tsLMs were first estimated on each time series independently, and then DTW distances were calculated to determine the neighbours of the target series. The forecast of the target series is generated by averaging local forecasts of its neighbours. The cluster-based models ignore the global information across the entire dataset. Meanwhile, clustering based on data similarity does not consider the model used, and it may divide the series that could be modelled using one tsGM into two clusters, leading to the loss of information and an increase in model complexity. 

\paragraph{Local-global hybrid models.}
To take advantage of the strength of both tsLMs and tsGMs, a local-global hybrid time series modelling method is introduced. \citet{Semenoglou2021} averaged the forecasts obtained by a local Theta method and a tsGM. \citet{Semenoglou2021} used some outputs of the tsLM as inputs to feed into the tsGM.  \citet{smyl2020hybrid} proposed a hybrid method of the exponential smoothing and RNNs, that is, the exponential smoothing method was used to capture the components of level and seasonality of the individual series, and a global RNN was applied to model the remaining homogeneous parts after local characteristics were removed. However, heterogeneity is not solely attributable to differences in level and seasonality. Insufficient removal of heterogeneity results in data patterns not being fully modelled. There is also some work combining linear models and non-linear models parallelly or serially, see \citet{Hajirahimi2019} and \citet{Zhang2003}, for example. However, they are built on one single time series, instead of  multiple time series. 

\paragraph{Error correction methods.}
Error correction is a technique to improve forecasting accuracy by residual modelling. A correction procedure is implemented after building a forecasting model, during which the residuals are modelled recursively until they are white noises. \citet{Firmino2015} used ARIMA models to recursively correct the forecasts obtained by neural networks, and discussed additive error models and multiplicative error models. Subsequently, \citet{Silva2019} used linear models (e.g., ARIMA) first and then employed non-linear models (e.g. support vector regression and LSTM) to correct errors and improve their model's forecasting accuracy. Boosting methods sequentially train weak learners, with each learner correcting the errors of predecessors at each iteration \citep{Binder2014}. Although these methods have some similarities to our method in this study, the error correction in their methods is a recursive combination of a sequence of linear and non-linear models. Their aim was to correct errors and adjust forecasts. Nevertheless, the aim of the present study is to identify and model the homogeneity among multiple time series and then capture the heterogeneity of the time series.

\section{Two-stage modelling framework} \label{hybrid-method}
\subsection{Problem Formulation}\label{hybrid-method-1}
Given a set of $n$ time series, we denote $\{x_{i, t}\}$ as the $i$-th time series for $i \in \{1, 2, \cdots, n\}$, $t \in \{t_{_0}^{(i)}, t_{_0}^{(i)}+1, \cdots, t_{_0}^{(i)}+n_i-1\}$, and $t_{_0}^{(i)}$ is the start time point, and $n_i$ is the length of the $i$-th series. Given a look-back window $\boldsymbol{x}_{i, (t-q):(t-1)} = [x_{i, t-q}, \cdots, x_{i, t-1}]$, a tsGM, $x_{i, t} = g(\boldsymbol{x}_{i, (t-q):(t-1)}| \Theta_g) + \epsilon_{i, t}$,  is learnt on the entire set of the time series, where $\Theta_g$ denotes the vector of all parameters in the function $g(.)$ and $\epsilon_{i, t}$ is the residual.

Heterogeneity among multiple time series poses a challenge for a tsGM in capturing all respective patterns of individual time series, even if the model is sufficiently complex. If the residuals of a tsGM model are tested as white noise, they are assumed not autocorrelated.  The constructed tsGM is deemed to be statistically adequate to model the corresponding time series. Otherwise, the tsGM is statistically inadequate and the residuals needs further analysing and modelling. Let $n_h$ denote the number of time series whose residuals are not white noise. Denote the heterogeneity level of the set of time series by $R_h$, which is defined as the ratio of $n_h$ to the total number of time series in this dataset, as shown below,
\begin{equation}
	R_h = R_h | (\{x_{i, t}\}, g) = \frac{ n_h}{n}.
\end{equation}
The heterogeneity level of a set of time series is relevant to the modelling method. For a given set of time series $\{x_{i, t}\}$ and a tsGM model $g(.)$, the larger $R_h$, the stronger heterogeneity.

In our proposed method, the tasks of learning a tsGM and identifying homogeneity are included at stage one, and those of modelling the heterogeneity are performed at stage two. Aggregating the results from two stages, the final time series model is updated to
\begin{equation}
	x_{i, t} = f\left(\boldsymbol{x}_{i, (t-q):(t-1)}, g(\boldsymbol{x}_{i, (t-q):(t-1)};\Theta_g), \Theta_g | \Theta_f\right) + \epsilon_{i, t},
\end{equation}
where $\Theta_f$ is the parameter vector to be estimated at stage two. 

The loss function used in this study is the mean squared error (MSE), which has the form of 
\begin{equation}
	\ell({x}_{i, t}, \widehat{x}_{i, t}) = \frac{1}{N}\sum_{i, t}({x}_{i, t}-\widehat{x}_{i, t})^2, 
	\label{MSE}
\end{equation}
where $\widehat{x}_{i, t} $ is the forecast and $N$ is the total number of sample points.

The empirical loss obtained by Equation~(\ref{MSE}) over the training samples is denoted as $R_{\text{emp}}$ and the expected loss over the entire data distribution $\mathcal{D}$ is given by $R_{\text{exp}}=\mathbb{E}_{(\boldsymbol{x}_{i, (t-q):(t-1)}, {x}_{i, t}) \sim \mathcal{D}}[({x}_{i, t} - \widehat{x}_{i, t})^2]$. The generalisation error of the model, denoted by $\varepsilon$, is defined as $\varepsilon = |R_{\text{emp}}-R_{\text{exp}}|$ \citep{Berner2022}, and it can be approximated as the difference between the in-sample error and the out-of-sample error.

\subsection{Model construction}
\begin{figure}[ht]
	\centering
	\includegraphics[width=8.5cm]{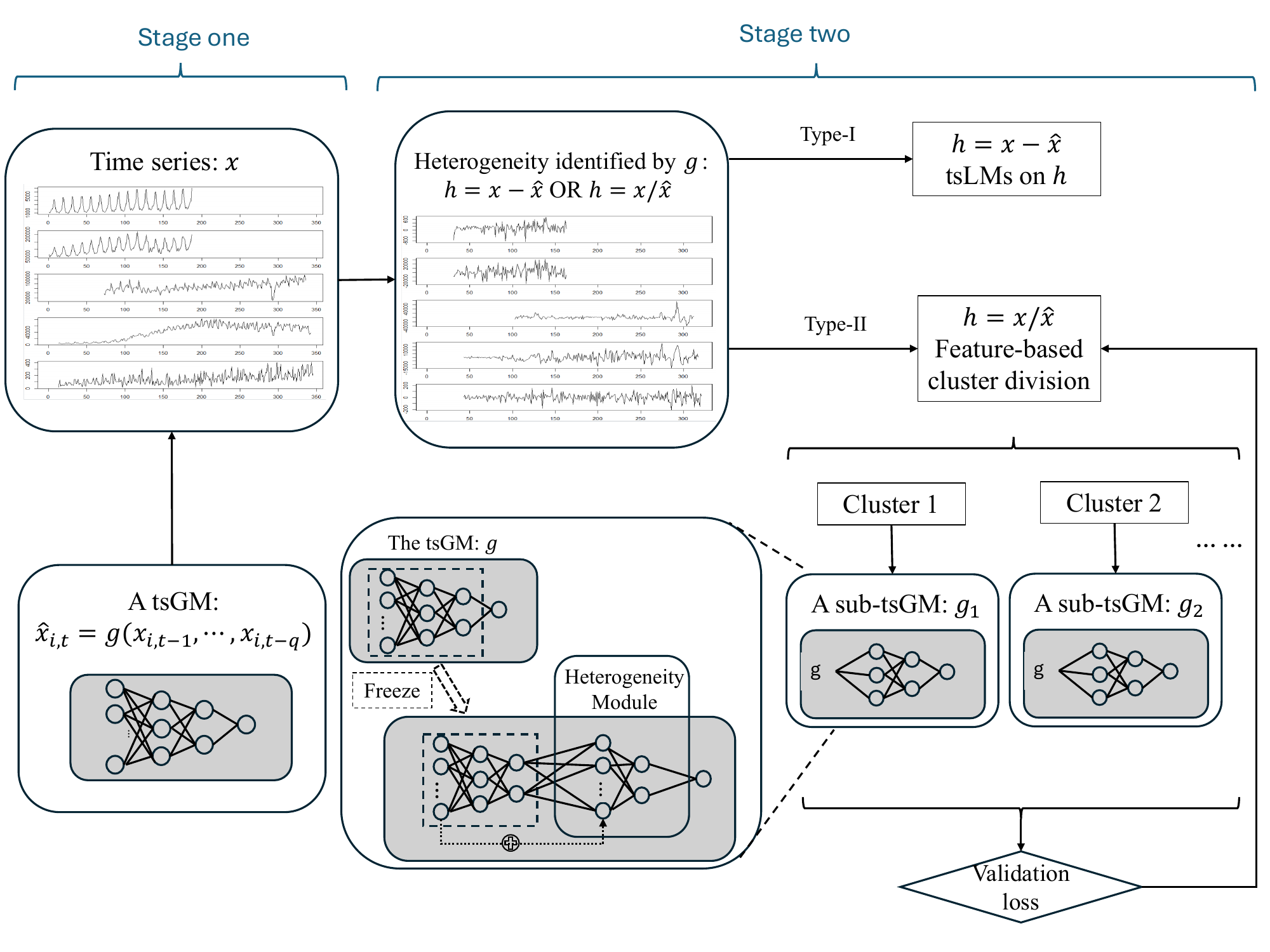}
	\caption{The two-stage modelling framework.}
	\label{fig:model_flow_chart}
\end{figure}

Figure~\ref{fig:model_flow_chart} illustrates the proposed two-stage modelling framework. At stage one, a tsGM $g(.)$ is learnt on the entire set of time series, based on which the model residuals can be calculated. The residuals are considered as including heterogeneity within this dataset, and heterogeneity series (denoted as $\{h_{i, t}\}$) can be described using either additive (i.e., $h_{i, t}=x_{i, t}-\widehat{x}_{i, t}$) or multiplicative (i.e., $h_{i, t}=x_{i, t}/\widehat{x}_{i, t}$) modelling methods. If the residuals of all individual time series are white noise as a result of some statistical tests (e.g., the Ljung-Box test), stage one is completed and stage two is not needed. Otherwise,  tsLMs and/or tsGMs must be estimated on the residuals at stage two to address the heterogeneity of the individual time series and extract remaining patterns. Two types of modelling methods are employed at stage two: Type-I and Type-II approaches. 

The Type-I approach constructs individual tsLMs such as the ARIMA model on each  heterogeneous residual series to supplement and update the forecasts generated by $g(.)$ at stage one. Theoretically, $R_h$ can be updated to 0. However, the costs of training and maintaining models dramatically increase when the number of heterogeneous time series is excessively large, and thus it is not applicable in this case. 
Alternatively, the Type-II approach estimates multiple sub-tsGMs for sub-groups of heterogeneous series, respectively. The identified heterogeneous series presents various autocorrelation, non-linearity or heteroskedasticity. Thus, feature-based clustering is considered. The residuals of the individual time series are grouped according to the features including autocorrelation coefficients, non-linearity (using the statistic of Teräsvirta’s non-linearity test \citep{terasvirta1993power}) and autoregressive conditional heteroskedasticity effects (using the statistic of Lagrange-multiplier test \citep{Engle1982}). Other features such as spectral entropy and lumpiness can also be involved \citep{hyndman2019tsfeatures}. These features are commonly-used and identified as features which influence the forecasting modelling and accuracy the most \citep{Talagala2022}.Through measuring the Euclidean distance between features, time series whose residual series exhibit similar features are put into a cluster. The number of clusters is denoted as $K$. A sub-tsGM, denoted by $g_k(.)$ ($k \in \{1, 2, \cdots, K\}$), is then trained on each cluster of time series. The structure of $g_k(.)$ is given in Figure~\ref{fig:model_flow_chart}, and Algorithm \ref{alg:g_k} presents the procedure of constructing it. After building all sub-tsGMs, an updated $R_h$ can be calculated. 

\SetKwComment{Comment}{/* }{ */}
\RestyleAlgo{ruled}
\begin{algorithm}[ht] 
	\caption{The procedure of constructing $g_k(.)$.}
	\label{alg:g_k}
	\KwIn{The tsGM at stage one $g(.)$ with $L$ layers}
	\KwOut{Cluster-specific sub-tsGM $g_k(.)$ with $L_k$ layers}
	\BlankLine
	\For{$l = 1$ \KwTo $L-1$}{
		Copy weights from layer $l$ of $g(.)$ to $g_k(.)$ \\
		Freeze layer $l$ in $g_k(.)$
	}
	Let $\mathbf{h}_{\text{global}} \leftarrow$ output of the first $L-1$ layer of $g_k(.)$\Comment*[r]{\scriptsize{Capture global features shared by the entire dataset}}
	Let $\mathbf{x}_{\text{input}} \leftarrow$ input of $g(.)$; \\
	Let $g_k^{(l)}(.) \leftarrow$ output of the $l$-th layer of $g_k(.)$; \\
	
	\BlankLine
	\textbf{Construct the heterogeneity module:} \\
	
	$\mathbf{z}_1 \leftarrow g_k^{(L)}(\mathbf{h}_{\text{global}})$;
	
	$\mathbf{z}_2 \leftarrow \text{reshape}(\mathbf{z}_1) + \mathbf{x}_{\text{input}}$\Comment*[r]{\scriptsize{Incorporate cluster-specific information}}
	
	$\widehat{x}_{i, t} \leftarrow g_k^{(L+1:L_k)}(\mathbf{z}_2)$\Comment*[r]{\scriptsize{Add more layers to capture cluster-specific patterns and adjust forecasts}}
	
\end{algorithm}

We split the entire set  of time series into  training, validation, and test subsets, and consider the in-sample error, the generalisation error and the model maintenance cost to optimise the clustering. Let the clustering $C$ based on $m$ features $fea_1, \cdots, fea_m$ be denoted by $C(fea_1, \cdots, fea_m)=\{C_1, C_2, \cdots, C_K\}$. The objective of the clustering is
\begin{eqnarray}
	\min_{C(fea_1, \cdots, fea_m)} &&  \sum_{k=1}^K \sum_{i \in C_k} \sum_{t \in {\rm val}} ({x}_{i, t}-\widehat{x}_{i, t})^2, \label{eq:clustering} \\
	{\rm s.t.} &&\sum_{k=1}^K |C_k| = n_h,  \notag \\
	&& \sum_{k=1}^K f_{m}(|C_k|) \leq U_m. \notag
\end{eqnarray}
Based on the training dataset, the optimisation defined in Equation~(\ref{eq:clustering}) aims to find a clustering $C$ which minimises the square error on the validation set. $|C_k|$ is the number of series in cluster $C_k$. We set an upper limit of the model maintenance cost as $U_m$. In reality, it can be relevant to the data volume (or model complexity) and the function $f_m(\cdot)$ demonstrates the relationship.

The two-stage modelling framework is denoted as TS-X-I/II, where X represents the tsGM estimated at stage one and I/II indicates Type-I or Type-II used at stage two. The following propositions are derived on the in-sample error and the generalisation error.

\begin{proposition} \label{prop:1}
	Based on the tsGM trained at stage one, stage two further reduces the MSE loss defined in Equation~(\ref{MSE}) by descending the gradient in the function space.
\end{proposition}
\begin{proof}
	The tsGM constructed at stage one is denoted as $g(.)$. $\forall i, t$, the functional gradient of the MSE loss at $g(.)$ can be derived as 	
	\begin{equation}
		\frac{\partial \ell({x}_{i, t}, g(\boldsymbol{x}_{i, (t-q):(t-1)}))}{\partial g(\boldsymbol{x}_{i, (t-q):(t-1)})} \propto {x}_{i, t} - g(\boldsymbol{x}_{i, (t-q):(t-1)}),
	\end{equation}
	which is proportional to the heterogeneity (i.e., residuals). Thus, for additive modelling,  adding a model describing heterogeneity at stage two is equivalent to conducting functional gradient descent in the function space. The MSE loss further decreases.
	
	The ${x}_{i, t}$ and $g(.)$ can always be shifted to positive by adding a fixed small value and then shifted back. Thus, in the log-space,
	\begin{equation}
		\frac{\partial \ell(\log({x}_{i, t}), \log\left(g(\boldsymbol{x}_{i, (t-q):(t-1)}))\right)}{\partial \log\left(g(\boldsymbol{x}_{i, (t-q):(t-1)})\right)} \propto \log\left(\frac{{x}_{i, t}}{g(\boldsymbol{x}_{i, (t-q):(t-1)})}\right).
	\end{equation}
	For multiplicative modelling, multiplying a model describing heterogeneity is equivalent to conducting functional gradient descent in the log-space. $\forall x_{i,t}$, the loss in the log-space, namely $\left(\log\left(\frac{{x}_{i, t}}{g(\boldsymbol{x}_{i, (t-q):(t-1)})}\right)\right)^2$, decreases, and the MSE loss decreases as well.
\end{proof}

Proposition \ref{prop:1} proves that stage two is conducive to reducing the in-sample error. The hypothesis set of the tsGM, which consists of all possible functions, is denoted as $\mathcal{H}$. The size of $\mathcal{H}$, which is the number of functions in $\mathcal{H}$ if $\mathcal{H}$ is finite, denoted as $|\mathcal{H}|$ \citep{Berner2022}. The size of hypothesis set of the model with trainable parameters learnt at stage two for cluster $k$ is denoted as $|\mathcal{H}_k|$, $k \in \{1, 2, \cdots, K\}$, respectively. $|\Phi|$ represents the size of the $k$-means clustering function class used at stage two and it is of order $\mathcal{O}(Km)$ \citep{pmlr-v139-li21k, shalev2014understanding}. Proposition \ref{prop:2} provides a generalisation upper bound of the proposed two-stage modelling framework.

\begin{proposition} \label{prop:2}
	If the identified heterogeneity is divided into $K$ clusters at stage two, with probability of at least $1-\delta$, it holds that
	\begin{equation}
	\varepsilon \leq \sqrt{\frac{\log(|\mathcal{H}|)+\log(\frac{4}{\delta})}{2\sum_{i=1}^n n_i^\prime}} + \sqrt{\frac{\log(|\Phi|)+\sum_{k=1}^{K}\log(|\mathcal{H}_k|)+\log(\frac{4}{\delta})}{2\sum_{i \in I_h} n_{i(h)}^\prime}},
	\notag
	\end{equation}	
	where $\delta \in (0, 1)$, and $n_i^\prime$ and $n_{i(h)}^\prime$ are the effective sample sizes \footnote{The effective sample size $n_i^\prime$ can be calculated using $n_i^\prime=n_i\left[\sum_{j=-(n_i-1)}^{+(n_i-1)}(1-\frac{|j|}{n_i})\rho(j)\right]^{-1}$, where $\rho(.)$ is the autocorrelation function \citep{Thiebaux1984}. And $n_{i(h)}^\prime$ can be obtained similarly.} of the $i$-th original and heterogneity time series, respectively, and $I_h$ is a set that contains the indices of all identified heterogeneity series.
\end{proposition}

\begin{proof}
	For Type-I modelling at stage two, $K$ can be regarded as equal to $n_h$, and $\mathcal{H}_k$ refers to the hypothesis set of each tsLM (e.g., ARIMA) in this case. While for the Type-II modelling method, $\mathcal{H}_k$ is the hypothesis set of the heterogeneity module constructed for cluster $k$. The MSE loss on normalised data can be regarded as bounded, and they can take values in [0, 1] after rescaling \citep{montero2021principles}. Following Theorem 1.17 of \citet{Berner2022}, with probability of at least $1-\delta/2$, the generalisation upper bound of each stage can be derived. A union bound of two stage is obtained and $P(\text{The upper bound of either stage fails}) \leq \delta/2 + \delta/2 = \delta$. The observations are sliding windows of time series and are non-i.i.d. The sample size is therefore replaced by the effective sample size \citep{montero2021principles}. The proposition is established.
\end{proof}

\begin{proposition} \label{prop:3}
	There exists an optimal clustering $C$ at stage two that minimises the out-of-sample error.
\end{proposition}
\begin{proof}
	Assuming the tsGM constructed at stage one identifies $n_h$ heterogeneous time series, the possible number of clusters at stage two ranges from 1 to $n_h$. Based on Propositions \ref{prop:1} and \ref{prop:2}, with well-fitted sub-tsGMs, an increase in the number of clusters reduces the in-sample error while increasing the generalisation error. Take two extreme scenarios as examples. If $K=n_h$, a Type-I modelling method at stage two is implemented and $n_h$ tsLMs are trained on each heterogeneous series, respectively. In-sample patterns can be completely captured as the residuals are checked as white noises. However, the risk of overfitting increases. If $K=1$, a single tsGM is trained at stage two. All heterogeneous series are described by one single model, and the underlying heterogeneity among these series obscures the modelling ability of capturing series-specific patterns. The effectiveness of the tsGM at stage two is limited to further improve the in-sample forecasting accuracy. However, the generalisation capability of tsGMs is fully utilised. Therefore, there must exist a $K$ $(1\leq K \leq n_h)$ that achieves a trade-off between in-sample error and generalisation error and minimises the out-of-sample error.
\end{proof}

Proposition \ref{prop:1} proves that modelling heterogeneity involves a one-step gradient descent in the loss function space. In fact, if we add more stages to implement clustering further, there always exists an equivalent clustering at stage two. Thus, the multi-stage setting is redundant and two stages suffice to approximate the target function. The key is to appropriately choose the features and the number of clusters and construct sub-tsGMs in light of out-of-sample error.

\section{Experiments and Results} \label{experiements}
\subsection{Experimental Setup}

\paragraph{Datasets.}
Four open datasets are used to evaluate the two-stage modelling framework. \textbf{Tourism}: It is from the tourism forecasting competition and is publicly available through the \textit{Tcomp} R package \citep{ellis2018tcomp}. \textbf{M3}: It has been used in the M3 forecasting competition and is provided in the R package \textit{Mcomp} \citep{hyndman2018package}. It is divided into five subcategories: M3-demographic, M3-finance, M3-industry, M3-macro, and M3-micro. The models are constructed on each subcategory separately and aggregated results are reported. \textbf{CIF 2016}: It is from the CIF 2016 forecasting competition, including 24 real-world series from the banking cluster and 48 artificially generated series. The dataset can be obtained from \citet{Neubauer2024}. \textbf{Hospital}: It tracks the number of patients for various medical problems and is publicly available from R package \textit{expsmooth} \citep{Hyndman2015}. 

The sampling frequency of these datasets is monthly and the statistical description is summarised in Table \ref{tab:datasets}, including the domain, number of series (\#Series), series length, mean trend and seasonality strength and Augmented Dick-Fuller (ADF) test statistics \citep{Elliott1996}. The trend and seasonality strength are calculated based on an STL decomposition \citep{hyndman2019tsfeatures}, and the value close to 1 indicates strong trend or seasonality. A smaller ADF test statistics indicates a more stationary time series.

\begin{table}[htbp]
	\centering
	\caption{The statistical description of datasets.}
	\setlength{\tabcolsep}{0.5mm}{
	\begin{tabular}{lrrrrrrcc}
		\toprule
		\multicolumn{1}{c}{\multirow{2}[0]{*}{Dataset}} & \multicolumn{1}{c}{\multirow{2}[0]{*}{Domain}} & \multicolumn{1}{c}{\multirow{2}[0]{*}{\#Series}} & \multicolumn{3}{c}{Length} & \multicolumn{1}{c}{\multirow{2}[0]{*}{Trend}} & \multicolumn{1}{c}{\multirow{2}[0]{*}{Seasonality}} & \multicolumn{1}{c}{\multirow{2}[0]{*}{ADF}} \\ \cline{4-6}
		&       &             & \multicolumn{1}{c}{min} & \multicolumn{1}{c}{max} & \multicolumn{1}{c}{mean} &       &       &  \\ 
		\midrule
		Tourism & \multicolumn{1}{c}{Tourism} & 366   & 91    & 333   & 299   & 0.855 & 0.752 & -5.788 \\
		Hospital & \multicolumn{1}{c}{Health care}  & 767   & 84    & 84    & 84    & 0.484 & 0.343 & -3.385 \\
		CIF 2016 & \multicolumn{1}{c}{Banking}  & 72    & 28    & 120   & 99    & 0.903 & 0.480 & -2.633 \\
		M3-demo & \multicolumn{1}{c}{Population}  & 111   & 71    & 138   & 123   & 0.944 & 0.348 & -2.738 \\
		M3-finance & \multicolumn{1}{c}{Markets}  & 145   & 68    & 144   & 124   & 0.906 & 0.331 & -1.925 \\
		M3-industry & \multicolumn{1}{c}{Manufacturing}  & 334   & 96    & 144   & 140   & 0.790 & 0.531 & -3.538 \\
		M3-macro & \multicolumn{1}{c}{Economy}  & 312   & 66    & 144   & 131   & 0.947 & 0.279 & -2.336 \\
		M3-micro & \multicolumn{1}{c}{Business}  & 474   & 68    & 126   & 93    & 0.493 & 0.371 & -4.063 \\
		M3    &    \multicolumn{1}{c}{-}   & 1,376  & 66    & 144   & 119   & 0.748 & 0.383 & -3.212 \\ \bottomrule
	\end{tabular}%
}
	\label{tab:datasets}%
\end{table}%

The time series in these datasets present characteristics of non-stationarity, including trends and seasonal patterns of different strengths. The problem of distribution shift affects the predictability of time series. \citet{kim2021reversible} proposed reversible instance normalisation (RevIN) to forecast non-stationary time series, and it applied normalisation with learnable parameters to each individual time series, and restored the statistical information of mean and variance on the corresponding outputs. Subsequently, \citet{liu2022non} experimentally found that this normalisation-and-denormalisation method is also effective without learnable parameters and named this revised design as {\it series stationarisation}. We apply series stationarisation to perform data pre-processing and post-processing. Concretely, normalisation is carried out on each sliding window over the temporal dimension. For  instance, the normalisation for an individual input $\boldsymbol{x}$ can be formulated as $\boldsymbol{x}^{\prime} = \frac{\boldsymbol{x} - \mu_x}{\sigma_x}$, where $\mu_x = \frac{1}{q}\sum_{j=1}^{q}x_{i, t-j}$ and $\sigma_x = \frac{1}{q}\sum_{j=1}^{q} (x_{i, t-j}-\mu_x)^2$. Suppose the corresponding forecasts $\widehat{x}_{i, t}^\prime$ are obtained through inputting instance $\boldsymbol{x}^\prime$ into the constructed model. Then, the denormalisation process transforms $\widehat{x}_{i, t}^\prime$ into the eventual forecasting results using $\widehat{x}_{i, t} = \widehat{x}_{i, t}^\prime \cdot \sigma_x + \mu_x$.

\paragraph{Baselines.}
The following baseline models are considered for model comparison.
\begin{itemize}
	\item \textbf{TSAVG}: It was proposed by \citet{Neubauer2024} in 2024. The $k$-nearest neighbour algorithm with DTW distances was utilised to form a neighbourhood of each time series, and the simple exponential smoothing is applied on each individual time series and the forecast is improved by averaging within its neighbourhood. Different averaging approaches include simple average, distance-weighted average and error-weighted average.
	\item \textbf{ARIMA}: Local ARIMA models are estimated on each individual time series \citep{Shumway2000} using \textit{auto.arima}() in the R package \textit{forecast} \citep{Hyndman2008}.
	\item \textbf{Pooled AR}: A pooled AR($q$) model, formulated as $x_{i, t} = \beta_0 + \beta_1 x_{i, t-1} + \cdots + \beta_q x_{i, t-q} + \epsilon_{i, t}$, is constructed on the entire dataset as a tsGM, where $\beta_0, \beta_1, \cdots, \beta_q$ are fitted by ordinary least squares. The order $q$ is determined following the practice of \citet{Neubauer2024}.
	\item \textbf{Multilayer Perceptron (MLP)}: It gains popularity due to its ability to model non-linear relationships \citep{Etemadi2023}. We consider dense layers, and the used activation function is the tanh function.
	\item \textbf{LSTM}: It was introduced by \citet{hochreiter1997long} in response to the problem of long-term dependencies. We consider LSTM layers followed by dropout layers to avoid overfitting.	
	\item \textbf{LSTM.Cluster}: \citet{Bandara2020} proposed the cluster-and-then-model methodology. Feature-based clustering is performed via the $k$-means algorithm and an LSTM network is constructed per cluster. The optimal number of clusters is determined by evaluating the validation loss.
\end{itemize}

\paragraph{Evaluation metrics.}
We focus on one-step-ahead forecasting and use cumulative errors as evaluation metrics \citep{Neubauer2024}. The cumulative one-step-ahead root mean squared error (RMSE), mean absolute error (MAE), and symmetric mean absolute percentage error (sMAPE) for the $i$-th time series have the following forms:
$
{\rm RMSE} = \frac{1}{n_{test}^{(i)}}\sum_{\tau=1}^{n_{test}^{(i)}}\sqrt{\frac{1}{\tau}\sum_{t=T}^{T+\tau-1}(x_{i, t}-\widehat{x}_{i,t})^2},
$
$
{\rm MAE} =\frac{1}{n_{test}^{(i)}}\sum_{\tau=1}^{n_{test}^{(i)}}\frac{1}{\tau}\sum_{t=T}^{T+\tau-1}|x_{i, t}-\widehat{x}_{i,t}|,
$
and
$
{\rm sMAPE} = \frac{1}{n_{test}^{(i)}}\sum_{\tau=1}^{n_{test}^{(i)}}\frac{2}{\tau}\sum_{t=T}^{T+\tau-1} \frac{|x_{i, t}-\widehat{x}_{i,t}|}{|x_{i, t}|+|\widehat{x}_{i,t}|},
$
respectively, where $n_{test}^{(i)}$ is the length of the test subset of the $i$-th time series, $i \in \{1, 2, \cdots, n\}$, and $T$ is the forecasting origin.

\subsection{Experimental Results}
The in-sample and out-of-sample data, divided by the dataset itself, are used as the training and test datasets, respectively. For the TSAVG, ARIMA and Pooled AR models, the experiments are implemented consistently with the settings of \citet{Neubauer2024}. While building MLP and LSTM neural networks, a validation dataset is needed to avoid overfitting and select the optimal hyperparameters. One subset of the training dataset is split as the validation dataset and the length is set as 10\% of the training dataset. Time series cross-validation with the one-step-ahead rolling origin setup is performed \citep{Hewamalage2022}. The learning rate is set to be 0.002 at the beginning and it is automatically updated using LearningRateScheduler with the rule of $learning \ rate*0.5^{epoch-1}$, and the number of epochs is 100 and early stopping is set based on the validation loss. Grid search is used to select the optimal hyperparameters and the ranges where the hyperparameters are selected are listed as follows: the input length $\in \{12, 24\}$, the number of layers $\in \{1, 2\}$, the number of nodes $\in \{4, 8, 16\}$, the dropout rate $\in \{0.2, 0.5\}$, and the batch size $\in \{32, 64\}$. The Adam optimiser is used~\citep{Kingma2014}. All experiments are conducted on Google Colab using a GPU-enabled runtime. The primary model computations are executed on an NVIDIA Tesla T4 GPU (16 GB VRAM). The environment also included an Intel Xeon CPU (2 cores, 2.20GHz) and 12 GB RAM running Ubuntu 22.04.4. 

The proposed two-stage models are developed based on the MLP and LSTM networks built above, namely, the MLP and LSTM models are the tsGM at stage one. The Ljung-Box test is performed, and the residual series with the $p$-value smaller than 0.05 indicate heterogeneity. At stage two, as shown in  Figure~\ref{fig:model_flow_chart}, there are two options. The Type-I is to construct all tsLMs on non-white noise residuals, we use \textit{auto.arima}() from the R package \textit{forecast} to build ARIMA models. For the Type-II   modelling method at stage two, the R package \textit{tsfeatures} is utilised to extract the features of residual series, including acf\_features, pacf\_features, entropy, lumpiness, stl\_features, arch\_stat, nonlinearity, unitroot\_kpss, unitroot\_pp, holt\_parameters, and hw\_parameters, and one may refer to \citet{hyndman2019tsfeatures} for the meaning of each feature, and we then carry out $k$-means algorithm to divide the residual series into different clusters. We simplify the maintenance cost in Equation~(\ref{eq:clustering}) to relate only to the number of models and ensure that the number of clusters does not exceed 10. After freezing weights and structure of the tsGM estimated at stage one, one more dense layer is added to it and the batch size at stage two decreases to 16, 8 or 4. We consider additive residuals for Type-I modelling and multiplicative residuals for Type-II according to experimental results. The forecasting performance is evaluated on the test dataset and the mean and median errors for each metric are calculated (see Table~\ref{tab:Model_Comparison} \footnote{The code is available at https://github.com/R-jr-star/Two-stage-modelling-framework.}), where the best-performed ones are in bold and the second-best ones are underlined. The tsGMs, especially neural networks, present flexible forecasting and generalisation abilities and outperform tsLMs such as the ARIMA. The proposed models significantly outperform the TSAVG, which is designed specifically for heterogeneous time series forecasting. Compared with the pure tsGMs, the TS-X-I/II is able to effectively identify and model heterogeneity, and further boost the forecasting performance.

\begin{table*}[ht] \centering 
	\caption{Model comparison among proposed models and baselines in terms of cumulative RMSE, MAE and sMAPE.}
	\setlength{\tabcolsep}{0.6mm}{
		\begin{tabular}{ccl|rrrrrr|rrrr}
			\toprule
			&       &       & \multicolumn{1}{c}{TSAVG} & \multicolumn{1}{c}{ARIMA} & \multicolumn{1}{c}{Pooled AR} & \multicolumn{1}{c}{MLP} & \multicolumn{1}{c}{LSTM} & \multicolumn{1}{c|}{LSTM.Cluster} & \multicolumn{1}{c}{TS-MLP-I} & \multicolumn{1}{c}{TS-LSTM-I} & \multicolumn{1}{c}{TS-MLP-II} & \multicolumn{1}{c}{TS-LSTM-II}  \\
			\midrule
			\multirow{6}[0]{*}{\rotatebox{90}{Tourism}} & \multirow{2}[0]{*}{RMSE} & mean  & 4759.500 & 4119.036 & 2047.617 & 1913.457 & 1963.280 &2192.981
			& \underline{1898.218} & 1903.357 & \textbf{1892.543} & 1916.188 \\
			&       & median & 1005.795 & 929.144 & 597.842 & 511.050 & 495.335 &509.936
			& 503.737 &\textbf{483.842} & 499.367 & \underline{490.785} \\
			& \multirow{2}[0]{*}{MAE} & mean  & 3647.229 & 3190.488 & 1543.199 & 1539.330 & 1540.164 &1809.114
			& 1529.742 & 1536.206 & \textbf{1513.960} & \underline{1515.407} \\
			&       & median & 770.753 & 700.724 & 461.658 & 393.885 & 384.459 & 418.610
			& 386.049 & 387.552 & \textbf{375.761} & \underline{383.539} \\
			& \multirow{2}[0]{*}{sMAPE} & mean  & 0.291 & 0.267 & 0.206 & 0.171 & \textbf{0.167} &0.185
			& 0.170 & 0.167 & 0.170 & \underline{0.167} \\
			&       & median & 0.262 & 0.241 & 0.158 & 0.136 & 0.136 &0.153
			& 0.138 & 0.142 & \underline{0.136} & \textbf{0.135} \\ \midrule
			
			\multirow{6}[0]{*}{\rotatebox{90}{M3}} & \multirow{2}[0]{*}{RMSE} & mean  & 615.987 & 598.826 & 597.438 & 562.183 & 579.985 &\textbf{557.876}
			& \underline{558.386} & 560.449 & 561.264 & 565.051 \\
			&       & median & 406.703 & 395.709 & 389.837 & \underline{328.168} & 358.744 & 356.812
			& \textbf{325.942} & 337.669 & 329.176 & 338.711 \\
			& \multirow{2}[0]{*}{MAE} & mean  & 484.660 & 470.654 & 475.890 & 464.153 & 478.653 & 463.335
			& \textbf{460.971} &\underline{462.290} & 462.659 & 465.618 \\
			&       & median & 322.573 & 311.126 & 313.683 & \underline{270.498} & 297.641 &290.237
			& \textbf{268.922} & 282.301 & 271.838 & 281.160 \\
			& \multirow{2}[0]{*}{sMAPE} & mean  & 0.115 & 0.114 & 0.114 & 0.113 & 0.113 &\textbf{0.110}
			& 0.114 & 0.112 & 0.113 & \underline{0.111} \\
			&       & median & 0.069 & 0.065 & 0.066 & \underline{0.055} & 0.062 & 0.061
			& \textbf{0.054} & 0.057 & 0.055 & 0.058 \\ \midrule
			
			\multirow{6}[0]{*}{\rotatebox{90}{CIF 2016}} & \multirow{2}[0]{*}{RMSE} & mean  & 360462.500 & 301763.100 & 451571.800 & 293090.638 & 268568.973 &279579.044
			& 308556.696 & \underline{266999.874} & 281854.068 & \textbf{248097.484} \\
			&       & median & 93.481 & 96.834 & 32944.598 & 78.259 & 87.128 & 87.080
			& \textbf{52.036} & \underline{54.156} & 76.271 & 75.733 \\
			& \multirow{2}[0]{*}{MAE} & mean  & 301584.200 & 233306.100 & 411428.100 & 239905.587 & 215052.708 &227962.398
			& 256551.538 & \underline{207039.528} & 225569.258 & \textbf{199321.157} \\
			&       & median & 82.064 & 79.007 & 32944.310 & 67.905 & 72.229 &73.351
			& \textbf{44.982} & \underline{50.356} & 65.871 & 62.792 \\
			& \multirow{2}[0]{*}{sMAPE} & mean  & 0.107 & 0.101 & 1.338 & 0.095 & 0.103 & 0.101
			& \textbf{0.082} & \underline{0.083} & 0.089 & 0.090 \\
			&       & median & 0.084 & 0.080 & 1.592 & 0.071 & 0.086 & 0.082
			& \underline{0.057} & \textbf{0.056} & 0.071 & 0.069 \\ \midrule
			
			\multirow{6}[0]{*}{\rotatebox{90}{Hospital}} & \multirow{2}[0]{*}{RMSE} & mean  & 23.985 & 23.320 & 21.610 & \underline{20.674} & 22.056 & 22.887
			& \textbf{20.596} & 21.888 & 20.926 & 21.961 \\
			&       & median & 8.080 & 8.022 & 8.357 & 7.833 & \underline{7.725} &7.663
			& 7.825 &\textbf{7.725} & 7.944 & 7.753 \\
			& \multirow{2}[0]{*}{MAE} & mean  & 19.610 & 19.040 &\underline{17.643} & 17.651 & 18.739 & 19.408
			&\textbf{17.574} & 18.654 & 17.816 & 18.635 \\
			&       & median & \textbf{6.437} & 6.495 & 6.854 & 6.600 & 6.558 & \underline{6.492}
			& 6.616 & 6.546 & 6.665 & 6.612 \\
			& \multirow{2}[0]{*}{sMAPE} & mean  & \textbf{0.168} & 0.169 & 0.176 & 0.169 & \underline{0.169} & 0.170
			& 0.170 & 0.169 & 0.170 & 0.170 \\
			&       & median & 0.158 & 0.161 & 0.169 & \underline{0.154} & 0.154 & 0.159
			& 0.154 & \textbf{0.153} & 0.154 & 0.156 \\
			\bottomrule
		\end{tabular}
	}
	\label{tab:Model_Comparison}
\end{table*}

\begin{figure}[ht]
	\centering
	\includegraphics[width=7cm]{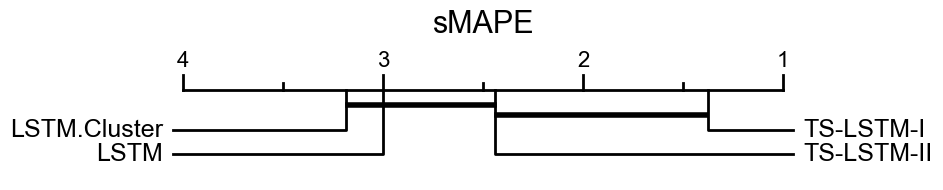}
	\caption{The CD diagram to visualise the differences among four LSTM-involved models in terms of mean sMAPE.}
	\label{fig:cd}
\end{figure}

The M3 dataset comprises five subsets, resulting in a total of eight datasets. Based on the obtained mean sMAPE, the Friedman test \citep{Friedman1940} is conducted to detect significant differences among the MLP, LSTM, LSTM.Cluster, and TS-MLP/LSTM-I/II models. The $p$-value is 0.096, indicating that there are significant differences between the performance of these models at a significance level of 0.1. The Nemenyi method \citep{nemenyi1963distribution} is employed as a post-hoc test to further evaluate four LSTM-involved models, and the Critical Distance (CD) diagram is plotted in Figure~\ref{fig:cd}. The TS-LSTM-I and TS-LSTM-II are not significantly different from each other while the TS-LSTM-I model performs significantly better than the LSTM and LSTM.Cluster. 

\section{Discussion}\label{discussion}
\paragraph{Heterogeneity level.}
Table~\ref{tab:table2} presents the heterogeneity levels detected by taking the MLP or the LSTM as the tsGM at stage one, and the residuals are calculated using the additive model. The number of series (\#Series), the number of sliding windows (\#Samples), and the number of model parameters (\#Parameters) are also listed. It can be found that even with the best-performed tsGM at stage one, the heterogeneity still exists and the level ranges from 9\% to 76\%. Stage two is necessary. Besides, we used the multiplicative model to obtain the heterogenous time series and applied the Type-II  modelling method at stage two. Table~\ref{tab:table3} summarises the heterogeneity levels before and after stage two. Adding stage two is conducive to reducing the heterogeneity level.

\begin{table}[ht]
	\centering
	\caption{The heterogeneity levels identified by tsGMs of the MLP and LSTM using the additive model.}
	\setlength{\tabcolsep}{1mm}{
		\begin{tabular}{llrrrrr} 
			\toprule
			\multicolumn{1}{c}{Dataset} & \makecell[c]{Global \\ model} & \multicolumn{1}{c}{\#Series} & \multicolumn{1}{c}{\#Samples} & \multicolumn{1}{c}{\#Parameters} & \multicolumn{1}{c}{$n_h$} & \multicolumn{1}{c}{$R_h$} \\
			\midrule
			\multirow{2}[0]{*}{Tourism} & MLP   & \multirow{2}[0]{*}{366} & \multirow{2}[0]{*}{91712} & 241   & 186   & 50.82\% \\
			& LSTM  &       &       & 1993  & 198   & 54.10\% \\ \midrule
			\multirow{2}[0]{*}{Hospital} & MLP   & \multirow{2}[0]{*}{767} & \multirow{2}[0]{*}{36816} & 417   & 92    & 11.99\% \\
			& LSTM  &       &       & 361   & 91    & 11.86\% \\ \midrule
			\multirow{2}[0]{*}{CIF 2016} & MLP   & \multirow{2}[0]{*}{72} & \multirow{2}[0]{*}{5470} & 113   & 34    & 47.22\% \\
			& LSTM  &       &       & 361   & 37    & 51.39\% \\ \midrule
			\multirow{2}[0]{*}{M3-demo} & MLP   & \multirow{2}[0]{*}{111} & \multirow{2}[0]{*}{9028} & 417   & 70    & 63.06\% \\
			& LSTM  &       &       & 1225  & 85    & 76.58\% \\ \midrule
			\multirow{2}[0]{*}{M3-finance} & MLP   & \multirow{2}[0]{*}{145} & \multirow{2}[0]{*}{11948} & 417   & 59    & 40.69\% \\
			& LSTM  &       &       & 1225  & 73    & 50.34\% \\ \midrule
			\multirow{2}[0]{*}{M3-industry} & MLP   & \multirow{2}[0]{*}{334} & \multirow{2}[0]{*}{32739} & 417   & 125   & 37.43\% \\
			& LSTM  &       &       & 1225  & 171   & 51.20\% \\ \midrule
			\multirow{2}[0]{*}{M3-macro} & MLP   & \multirow{2}[0]{*}{312} & \multirow{2}[0]{*}{27731} & 417   & 133   & 42.63\% \\
			& LSTM  &       &       & 1225  & 158   & 50.64\% \\ \midrule
			\multirow{2}[0]{*}{M3-micro} & MLP   & \multirow{2}[0]{*}{474} & \multirow{2}[0]{*}{24009} & 417   & 56    & 11.81\% \\
			& LSTM  &       &       & 1225  & 44    & 9.28\% \\
			\bottomrule
		\end{tabular}%
	}
	\label{tab:table2}%
\end{table}%

\begin{table}[ht]
	\centering
	\caption{The change of heterogeneity levels before and after the Type-II modelling method at stage two using the multiplicative model.}
	\setlength{\tabcolsep}{1mm}{
		\begin{tabular}{llrrrrrr}
			\toprule
			\multirow{2}{*}{Dataset} & \multirow{2}{*}{\makecell[c]{Global \\ model}} & \multicolumn{2}{c}{Before} &   &    & \multicolumn{2}{c}{After} \\ \cline{3-4} \cline{7-8}
			&  & \multicolumn{1}{c}{$n_h$} & \multicolumn{1}{c}{$R_h$} & $K$ & \multicolumn{1}{c}{\#Parameters} & \multicolumn{1}{c}{$n_h$} & \multicolumn{1}{c}{$R_h$} \\
			\midrule
			\multirow{2}[0]{*}{Tourism} & MLP   & 139   & 37.98\% & 5     & 173   & 64    & 17.49\% \\
			& LSTM  & 162   & 44.26\% & 5     & 173   & 74    & 20.22\% \\ \midrule
			\multirow{2}[0]{*}{Hospital} & MLP   & 81    & 10.56\% & 3     & 461   & 10    & 1.30\% \\
			& LSTM  & 88    & 11.47\% & 3     & 173   & 29    & 3.78\% \\ \midrule
			\multirow{2}[0]{*}{CIF 2016} & MLP   & 33    & 45.83\% & 2     & 137   & 16    & 22.22\% \\
			& LSTM  & 35    & 48.61\% & 2     & 89    & 14    & 19.44\% \\ \midrule
			\multirow{2}[0]{*}{M3-demo} & MLP   & 59    & 53.15\% & 3     & 461   & 34    & 30.63\% \\
			& LSTM  & 84    & 75.68\% & 3     & 173   & 52    & 46.85\% \\ \midrule
			\multirow{2}[0]{*}{M3-finance} & MLP   & 63    & 43.45\% & 3     & 461   & 18    & 12.41\% \\
			& LSTM  & 67    & 46.21\% & 3     & 173   & 31    & 21.38\% \\ \midrule
			\multirow{2}[0]{*}{M3-industry} & MLP   & 118   & 35.33\% & 5     & 461   & 46    & 13.77\% \\
			& LSTM  & 169   & 50.60\% & 3     & 173   & 77    & 23.05\% \\ \midrule
			\multirow{2}[0]{*}{M3-macro} & MLP   & 128   & 41.03\% & 3     & 461   & 53    & 16.99\% \\
			& LSTM  & 156   & 50.00\% & 3     & 173   & 64    & 20.51\% \\ \midrule
			\multirow{2}[0]{*}{M3-micro} & MLP   & 40    & 8.44\% & 3     & 461   & 14    & 2.95\% \\
			& LSTM  & 37    & 7.81\% & 3     & 173   & 16    & 3.38\% \\
			\bottomrule
		\end{tabular}%
	}
	\label{tab:table3}%
\end{table}%

\begin{figure}
	\centering
	\subfigure[]{\includegraphics[width=0.23\textwidth]{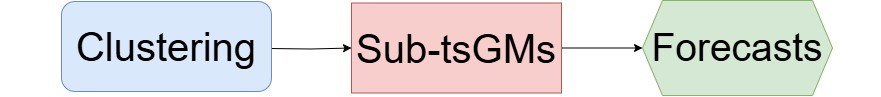} \label{fig:global_info_a}}
	\subfigure[]{\includegraphics[width=0.23\textwidth]{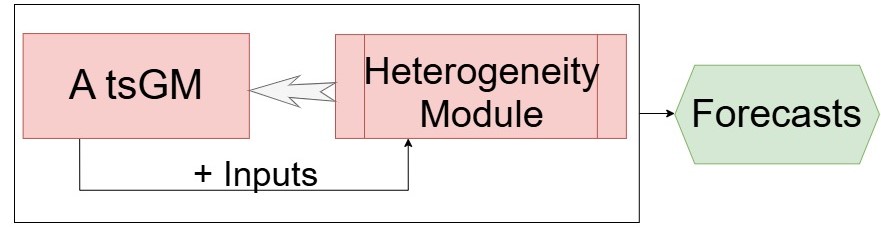} \label{fig:global_info_b}}
	\subfigure[]{\includegraphics[width=0.23\textwidth]{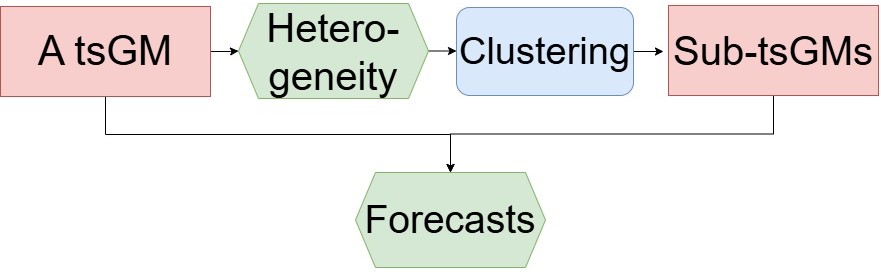} \label{fig:global_info_c}}
	\subfigure[]{\includegraphics[width=0.23\textwidth]{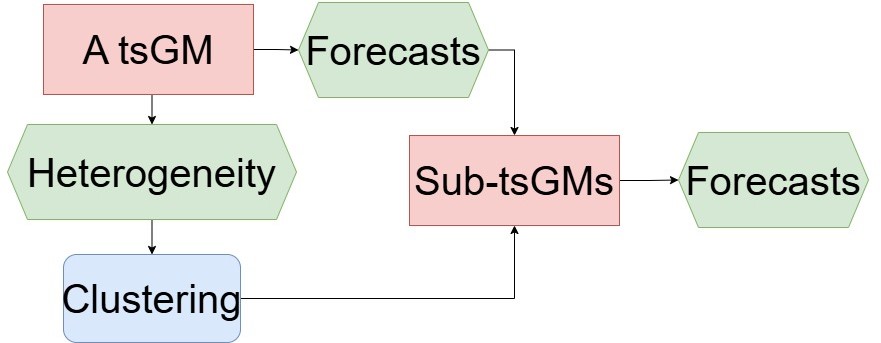} \label{fig:global_info_d}}
	\caption{Different strategies to address heterogeneity.}
	\label{fig:integrate_global_info}
\end{figure}

\paragraph{Significance of global information.}
One commonly-used strategy to address heterogeneity is clustering. As shown in Figure~\ref{fig:global_info_a}, a sub-tsGM is trained per cluster and the LSTM.Cluster falls within this category. In this way, however, only sub-global information within cluster is considered. Beyond the methodology proposed in this study (see Figure~\ref{fig:global_info_b}), there are several alternative approaches to incorporate global information. In Figure~\ref{fig:global_info_c}, heterogeneity is similarly identified by the developed tsGM. Rather than adding a heterogeneity module to the tsGM, the clustering-and-then-model method is directly applied to heterogeneity. In Figure~\ref{fig:global_info_d}, feature-based clustering is conducted on identified heterogeneity to form clusters. Within each cluster, the forecasts generated by the tsGM are concatenated with the corresponding time series data to serve as inputs for training a sub-tsGM, and the final forecasts are produced.

Table~\ref{tab:t_test} presents a comparative analysis of these different strategies based on the mean values of sMAPE. To ensure a fair comparison, the sub-tsGMs employed in strategies (c) and (d) are developed as LSTM networks as well, and hyperparameters and architectural configurations keep consistent with the setting of heterogeneity module of the TS-LSTM-II model. One-tail pairs $t$-tests are carried out to compare strategies (b)--(d) and (a) with the significance level of 0.1. It can be obtained from $p$-values that integrating global information of the entire dataset leads to statistically significant accuracy improvements, and the improvement demonstrated by the proposed TS-LSTM-II is the most significant.

\begin{table}[H]
	\centering
	\caption{Evaluation of various strategies for addressing heterogeneity in terms of mean values of cumulative sMAPE.}
	\setlength{\tabcolsep}{1.5mm}{
		\begin{tabular}{lcccc}
			\toprule
			& \multicolumn{1}{c}{(b) TS-LSTM-II} & \multicolumn{1}{c}{(c)} & \multicolumn{1}{c}{(d)} & \multicolumn{1}{c}{(a) LSTM.Cluster} \\ \midrule
			Tourism & 0.1672 & 0.1681 & 0.1681 & 0.1850 \\
			CIF 2016 & 0.0897 & 0.0843 & 0.0900 & 0.1018 \\
			Hospital & 0.1696 & 0.1691 & 0.1690 & 0.1698 \\
			M3-demo & 0.0414 & 0.0413 & 0.0421 & 0.0442 \\
			M3-finance & 0.0722 & 0.0712 & 0.0716 & 0.0755 \\
			M3-industry & 0.0851 & 0.0863 & 0.0854 & 0.0901 \\
			M3-macro & 0.0366 & 0.0375 & 0.0365 & 0.0377 \\
			M3-micro & 0.2069 & 0.2081 & 0.2073 & 0.1978 \\ \midrule
			$p$-value     & 0.094$^*$ & 0.102 & 0.097$^*$ &  \\ \bottomrule
		\end{tabular}%
	}
	\label{tab:t_test}%
\end{table}%

\begin{figure}
	\centering	
	\subfigure[RMSE]{\includegraphics[width=0.157\textwidth]{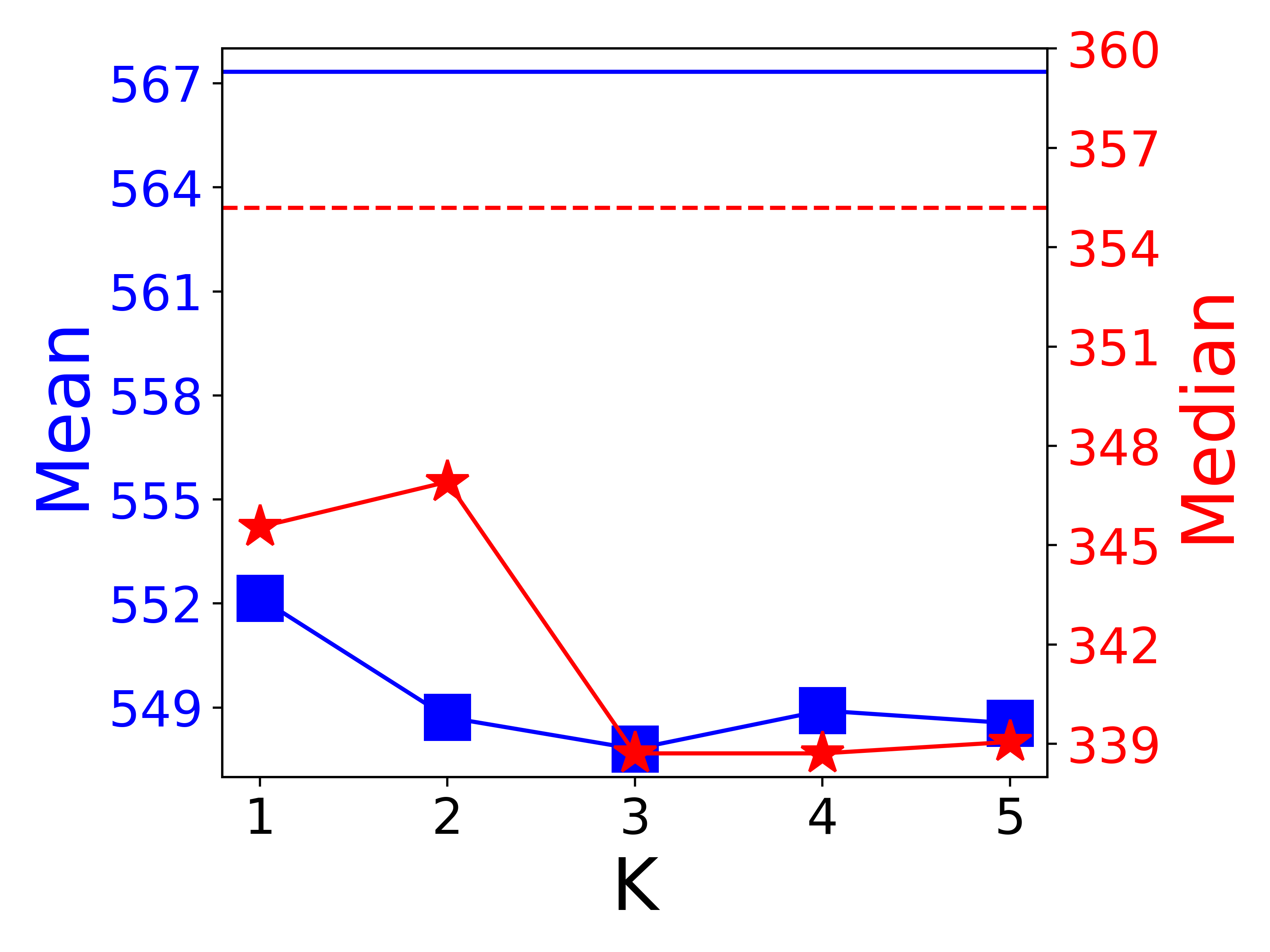}}
	\subfigure[MAE]{\includegraphics[width=0.157\textwidth]{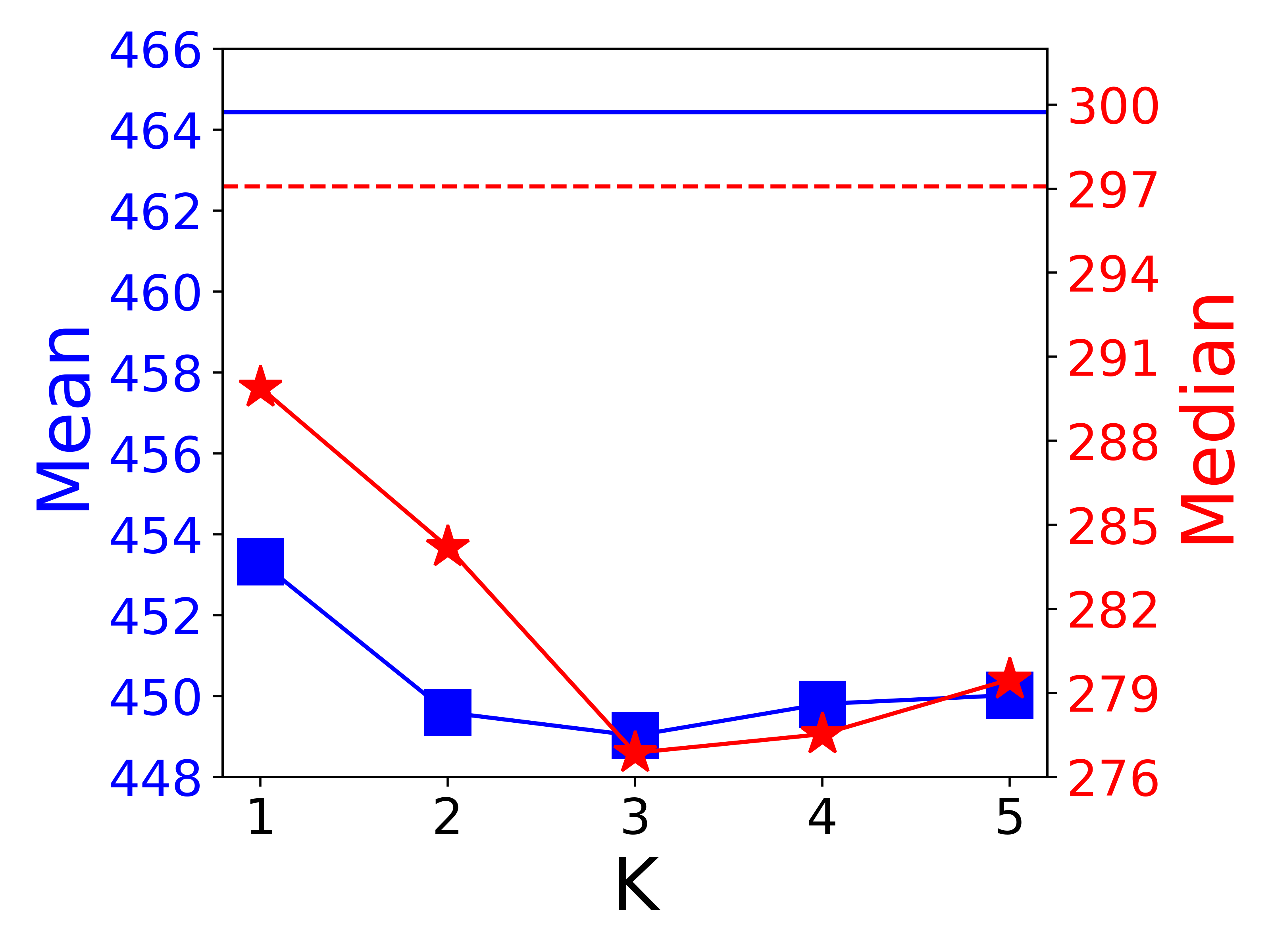}}
	\subfigure[sMAPE]{\includegraphics[width=0.157\textwidth]{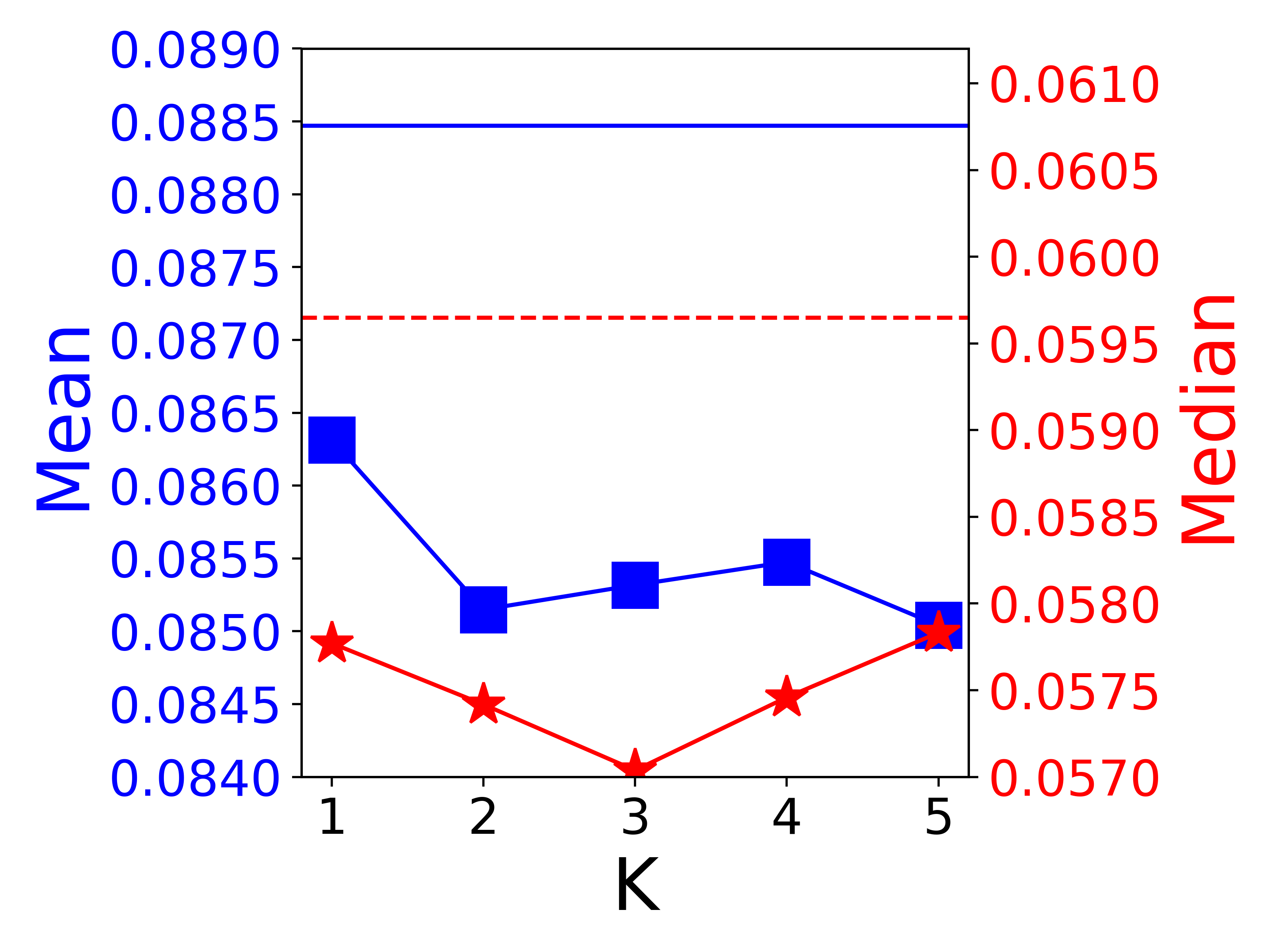}}
	\caption{Mean and median RMSE, MAE and sMAPE on M3-industry with different $K$ (Square markers and the solid horizontal line --  mean error).}
	\label{fig:sensitivity_k}
\end{figure}

\paragraph{Sensitivity analysis on $K$.}
The maintenance cost in Equation~\eqref{eq:clustering} constraints both $K$ and the number of series in each cluster. In practice, a form of $f_m(.)$ can be linear or nonlinear as different companies have various strategies \citep{sculley2015hidden}. The sensitivity of the number of clusters $K$ is analysed on the test dataset. M3-industry is taken as an example for illustration (see Figure \ref{fig:sensitivity_k}). The tsGM used at stage one is LSTM, and two horizontal lines are the values of corresponding test errors when only the first stage is considered. It further affirms Proposition \ref{prop:3} that there exists an optimal $K$ to realise a trade-off and minimise the out-of-sample error. Experimental results reveal the optimal number of clusters is generally between 2 to 5 (incl.).

\paragraph{Runtime analysis.} 
The results on runtime of LSTM-involved models across different datasets are summarised in Table \ref{tab:runtime}. The Type-I modelling at stage two is time-efficient. The TS-LSTM-II involves strenuous computation although it achieves more accurate forecasting. The increase in running time is primarily due to the small batch size employed at stage two to prevent overfitting. 

\begin{table}[H]
	\centering
	\caption{Runtime (in minutes) for each model across different datasets.}
	\begin{tabular}{lrrrr}
		\toprule
		& \multicolumn{1}{c}{LSTM} & \multicolumn{1}{c}{LSTM.Cluster} & \multicolumn{1}{c}{TS-LSTM-I} & \multicolumn{1}{c}{TS-LSTM-II} \\ \midrule
		Tourism & 10.75 & 10.95 & 11.20 & 22.24 \\
		Hospital & 4.17  & 6.82  & 4.66  & 7.94 \\
		CIF 2016 & 0.71  & 1.72  & 0.72  & 3.16 \\
		M3-demo & 1.11  & 2.85  & 1.55  & 7.98 \\
		M3-finance & 1.17  & 3.90  & 1.46  & 6.33 \\
		M3-industry & 3.54  & 10.24 & 4.18  & 18.40 \\
		M3-macro & 2.52  & 8.74  & 3.01  & 16.19 \\
		M3-micro & 2.35  & 8.26  & 2.47  & 4.50 \\ \bottomrule
	\end{tabular}%
	\label{tab:runtime}%
\end{table}%

\begin{figure}[H]
	\centering
	\subfigure[The 1-st time series]{\includegraphics[width=0.19\textwidth]{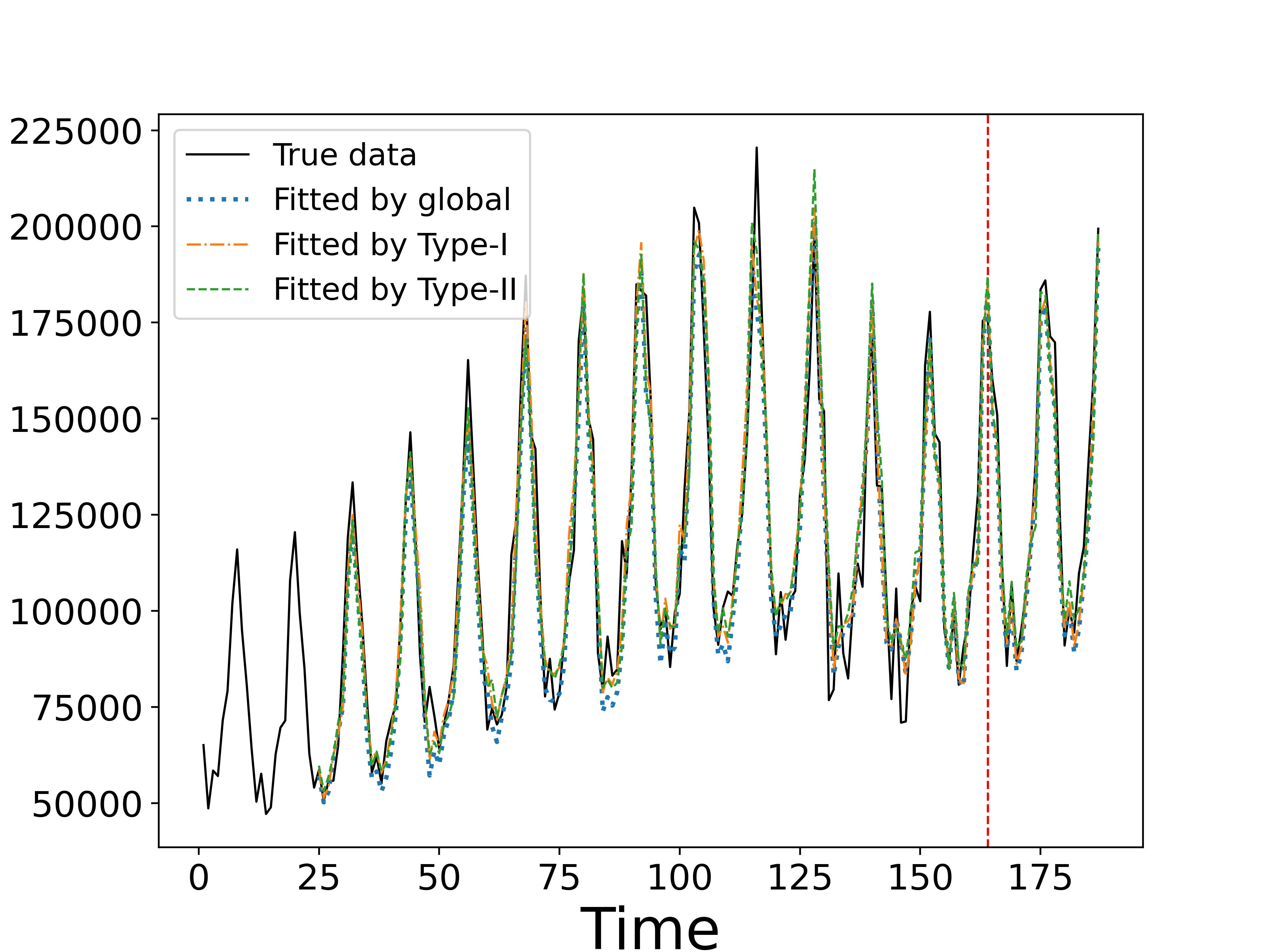}}
	\subfigure[The 112-nd time series]{\includegraphics[width=0.19\textwidth]{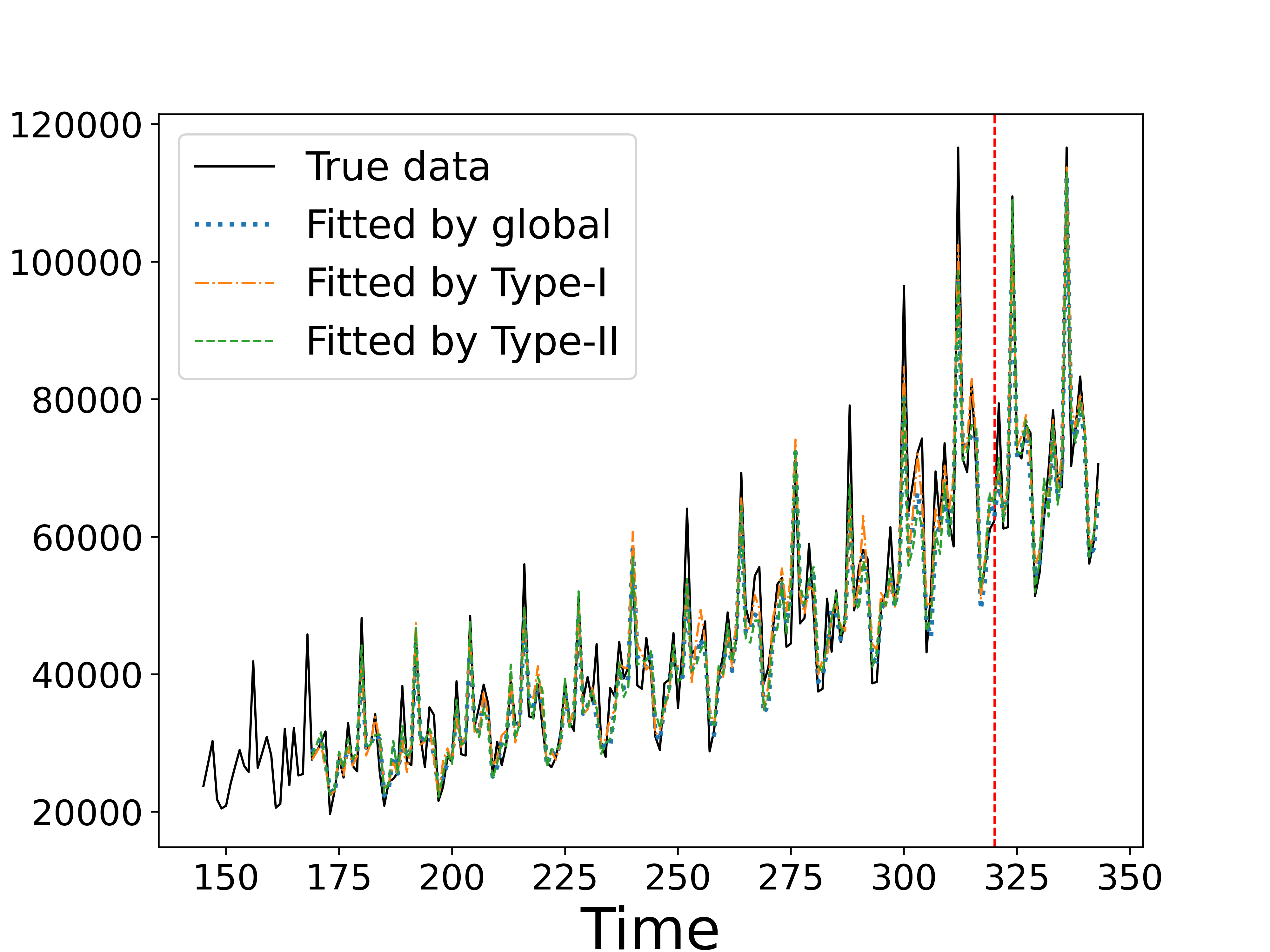}}
	\caption{The plots of true values and forecasts on the Tourism dataset, where the red vertical dashed line separates the training and the test datasets.}
	\label{fig:forecast}
\end{figure}

\paragraph{Visualisation of forecasting results.}
With the LSTM as the tsGM at stage one, the stage two leads to effective adjustments to generated forecasts (see Figure~\ref{fig:forecast}).

\section{Conclusion}\label{conclusion}
Considering characteristics of both time series and modelling methods, this paper proposed a two-stage modelling framework to boost global time series forecasting models on heterogeneous datasets. It can leverage local information of individual series, sub-global information within each sub-group, and global information across the entire dataset. This provides an alternative modelling approach to practitioners and researchers. Alternative techniques on data normalisation such as SAN \citep{liu2023adaptive} can be considered in future work. 


\newpage
\begin{ack}
	This work was supported by the Economic and Social Research Council of UK (ES/P00072X/1: 2617249, ESRC Standard Research Studentship: 22020946).
\end{ack}



\bibliography{ref}

\end{document}